\documentclass{article}
\usepackage{ijcai24}

\usepackage{times}
\usepackage{soul}
\usepackage{url}
\usepackage[hidelinks]{hyperref}
\usepackage[utf8]{inputenc}
\usepackage[small]{caption}
\usepackage{graphicx}
\usepackage{amsmath}
\usepackage{amsthm}
\usepackage{booktabs}
\usepackage{algorithm}
\usepackage{algorithmic}
\usepackage[switch]{lineno}

\usepackage{color}

\title{A Comprehensive Survey on 3D Content Generation}

\author{
Jian Liu$^{1,2}$\and
Xiaoshui Huang$^2$\thanks{Corresponding author}\and
Tianyu Huang$^1$\and
Lu Chen$^{2,4}$ \and
Yuenan Hou$^2$ \and
Shixiang Tang$^5$\\
Ziwei Liu$^3$ \and
Wanli Ouyang$^2$\and 
Wangmeng Zuo$^1$\and 
Junjun Jiang$^1$\and
Xianming Liu$^1$\footnotemark[1] \\
\affiliations
$^1$Harbin Institute of Technology, $^2$Shanghai AI Laboratory, $^3$S-Lab,Nanyang Technological University \\
$^4$Zhejiang University, $^5$The Chinese University of Hong Kong \\
\emails
\ ziwei.liu@ntu.edu.sg,
huangxiaoshui@pjlab.org.cn,
csxm@hit.edu.cn
}
\begin{document}
\maketitle

\begin{abstract}

Recent years have witnessed remarkable advances in artificial intelligence generated content (AIGC), with diverse input modalities, \emph{e.g.}, text, image, video, audio and 3D. The 3D is the most close visual modality to real-world 3D environment and carries enormous knowledge. The 3D content generation shows both academic and practical values while also presenting formidable technical challenges. This review aims to consolidate developments within the burgeoning domain of 3D content generation. Specifically, a new taxonomy is proposed that categorizes existing approaches into three types: 3D native generative methods, 2D prior-based 3D generative methods and hybrid 3D generative methods. The survey covers approximately \textbf{60} papers spanning the major techniques.
Besides, we discuss limitations of current 3D content generation techniques, and point out open challenges as well as promising directions for future work. Accompanied with this survey, we have established a project website where the resources on 3D content generation research are provided. The project page is available at \href{https://github.com/hitcslj/Awesome-AIGC-3D}{https://github.com/hitcslj/Awesome-AIGC-3D}.

\end{abstract}

\section{Introduction}

The generative models have gained tremendous success in natural language processing (NLP) \cite{achiam2023gpt} and image generation \cite{betker2023improving}. The recent developments, such as ChatGPT and Midjourney, have revolutionized many academic and industrial fields. For example, the artificial intellegence (AI) writing and designing assistants have remarkably shortened the duration of the paper writing and image design, respectively. In the 3D field, the 3D generation technologies have also made significant strides with the increasing amount of 3D data and the generation technology success of other fields.

The research of 3D content generation is attracting increasing interests due to its wide applications. The typical application is game and entertainment design \cite{liu2021deep}. The traditional design in game and entertainment, such as roles and objects, requires multiple views concept design, 3D model creation and 3D model refinement. This process is labour-intensive and time-consuming. The 3D content generation technology will largely reduce the time and labor cost. Another application is the construction field. With the 3D content generation methods, the designer can quickly generate the 3D concept models and communicate with the customer. This will narrow the gap between designer and customer, and will transform the construction design field. The third application is the industrial design \cite{liu20233dall}. The current industrial design requires the 3D part model generation, then assemble them into an integrated model. This process is time-consuming and may cost much material waste. The 3D content generation technology will produce all the 3D models virtually and assemble them into an integrated model. If the model is not satisfied, the designer can quickly revise the design without much cost.

The past several years have witnessed many advancements in the 3D native generative methods \cite{shi2022deep,li2023generative}. The main idea of these methods is to first train the network using 3D datasets and then generate 3D assets in a feed-forward manner. One limitation of this line of methods is the requirement of vast amount of 3D datasets while the amount of 3D data is scarce. Since the quantity of image-text pairs is much larger than the 3D counterparts, there recently emerges a new research direction by building 3D models upon 2D diffusion models that are trained on large-scale paired image-text datasets. One representative is DreamFusion ~\cite{poole2023dreamfusion}, which optimizes a NeRF by employing the score distillation sampling (SDS) loss. There recently also emerges hybrid 3D generative methods by combining the advantage of 3D native and 2D prior-based generative methods. The typical example is one2345++\cite{liu2023one2345plus} which generates 3D model by training a 3D diffusion model with input of 2D prior-based multi-view images. The recent two years have witnessed significant development in the 3D generative technologies, especially in text-to-3D~\cite{poole2023dreamfusion,lin2023magic3d} and image-to-3D~\cite{liu2023one2345plus,liu2023syncdreamer} tasks. These developments have provided many potential solutions in 3D content generation, such as 3D native generation, 2D prior-based 3D generation and hybrid 3D generation. 

Based on our best knowledge, there are only two surveys relevant to ours \cite{shi2022deep,li2023generative}. \cite{shi2022deep}  barely covers early techniques in shape generation and single view reconstruction.  \cite{li2023generative} only includes partial 2D prior-based 3D generative methods and not covers most recently 3D native and hybrid generative methods. However, this field has endured fast developments including 3D native, 2D prior-based and hybrid generative methods. Therefore, there is an urgent need for a comprehensive survey that consolidates these new developments and helps practitioners better navigate the expanding research frontier. 

In this survey, we make the following contributions. First, we propose a new taxonomy to systematically categorize the most recent advances in the 3D content generation field. Second, we provide a comprehensive review that covers 60 papers spanning the major techniques. Lastly, we discuss several promising future directions and open challenges. The survey caters to the requirement of both industrial and academic communities.

The paper is organized as follows: section 2 introduces preliminaries including 3D representation and diffusion models; section 3,4,5 introduce 3D generative methods covering 3D native, 2D prior-based, hybrid respectively. section 6 summarizes future directions in this promising AIGC-3D field. The last section concludes this survey.

\section{Preliminaries}
\label{Preliminaries}

\subsection{3D Representation}

Effectively representing 3D geometric data is crucial for generative 3D content. Introduce 3D representation is critical for understanding generative 3D content. The current 3D representation is typically classified into two categories, \emph{i.e.}, explicit and implicit representations. This section provides an overview of these representations.

\subsubsection{Explicit Representations}

Explicit representation usually refers to the direct and explicit representation of the geometry or structure of the 3D object. It involves explicitly defining the surface or volumetric representation of the object, such as through the use of point clouds, voxel grids or meshes. The advantage of explicit representation is that it enables more precise geometric control and multi-scale editing.

\noindent \textbf{Point Cloud}.
A point cloud is a fundamental representation for 3D data that involves sampling surface points from a 3D object or environment. Point clouds are often directly obtained from depth sensors, resulting in their widespread application in diverse 3D scene understanding problems. Depth maps and normal maps can be viewed as specific instances of the point cloud paradigm. Given the ease of acquiring point cloud data, this representation sees extensive usage in the domain of AIGC-3D.

\noindent \textbf{Voxel}.
The voxel is another common 3D representation that involves assigning values on a regular, grid-based volumetric structure. This allows a voxel to encode a 3D shape or scene. Due to the regular nature of voxels, they integrate well with convolutional neural networks and see extensive application in deep geometry learning tasks. Owing to this compatibility with CNNs, voxels are also a frequent choice for generative 3D content techniques that leverage deep neural models.

\noindent \textbf{Mesh}.
The mesh representation models 3D shapes and scenes using a collection of vertices, edges and faces. This allows meshes to encode both the 3D positional information and topological structure of surfaces. In contrast to voxels, meshes exclusively focus on modeling surface geometries, providing a more compact storage format. When compared to point clouds, meshes furnish explicit connectivity between surface elements, enabling the modeling of spatial relationships among points. Due to these advantages, meshes have long been widely used in classic computer graphics domains such as geometry processing, animation and rendering where accuracy, interoperability and efficiency are priorities. Striking a balance across these dimensions, meshes have emerged as a predominant representation in 3D content creation.

\subsubsection{Implicit Representations}

Implicit representation defines the 3D object or shape implicitly. A level set or a function that represents the object's surface is usually adopted. It offers a compact and flexible representation of 3D shapes, allowing for the modeling of objects, scenes, humans with sophisticated geometry and texture. The advantage of implicit representation lies in the flexible embedding with differential rendering pipeline.

\noindent \textbf{NeRF}.
Neural Radiance Fields (NeRF) is an emerging neural rendering method that has achieved impressive results for novel view synthesis of complex scenes.
NeRF consists of two primary components, including a volumetric ray tracer and a Multi-Layer Perceptron (MLP). NeRF is commonly used as a global representation in AIGC-3D applications, though it can be slow for rendering outputs.

\noindent \textbf{3D Gaussian Splatting}.
The 3D Gaussian Splatting (3D GS)~\cite{kerbl3Dgaussians} introduces an effective approach for novel view synthesis that represents 3D scenes implicitly using a set of weighted Gaussian distributions located in the 3D space. By modeling surface elements or points as Gaussian blobs, this method is able to capture complex scene structures with a sparse set of distributions. The ability to encode rich scene information implicitly via a distribution-based paradigm makes 3D Gaussian Splatting stand out as an innovative technique in novel view synthesis. 3D Gaussian Splatting has also recently seen application in AIGC-3D, though it produces results quickly but in an unstable manner.

\noindent \textbf{Signed Distance Function}.
The Signed Distance Function (SDF), defines a 3D surface as the zero level set of a distance field, where each point in space is assigned a value corresponding to its signed shortest distance to the surface. SDFs allow for efficient operations like construted solid geometry by utilizing distance values without requiring explicit mesh representations. They enable smooth surface reconstruction and support advanced simulations through level set methods. DMTet employs a hybrid representation combining Signed Distance Functions (SDFs) and meshes which is commonly used to refine and optimize generated 3D geometries.

\begin{figure*}[ht]
	\centering
	\includegraphics[width=\textwidth,trim=10bp 15bp 5bp 4bp,clip]{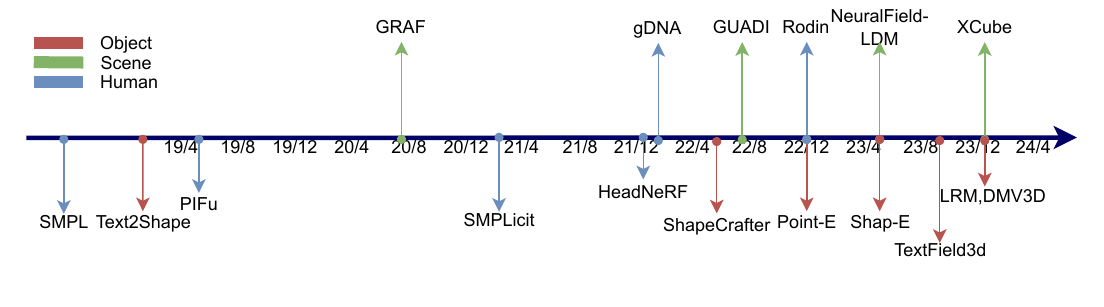}
	\caption{Chronological overview of the most relevant 3D native generative methods.
	}
	\label{fig:3dnative}
\end{figure*}

\subsection{2D Diffusion Models}

Diffusion models refer to a class of generative techniques based on the Denoising Diffusion Probabilistic Model (DDPM) framework. DDPM trains a model to perform the reverse diffusion process - starting from a noisy signal and applying iterative denoising steps to recover the original data distribution. Mathematically, this process can be represented as $x_t \sim p(x_t|x_{t-1})$, where $x_t$ is a noisy version of the original signal $x_0$ after $t$ diffusion steps with added Gaussian noise $\epsilon_t \sim \mathcal{N}(0,\sigma_t^2 I)$. By learning to denoise across different noise levels, the model can generate new samples by starting with random noise and applying the reverse diffusion process. 

\section{3D Native Generative Methods}

The 3D native generative methods directly generate 3D representations after the supervision of 3D data, in which representation and supervision are two crucial components for the generation quality. Existing 3D native generative methods can be classified three categories: object, scene and human. Several milestone methods
are presented in Figure~\ref{fig:3dnative}.

\subsection{Object}

With proper conditional input, 3D native generators can be trained for object-level generation. The early attempt, such as Text2Shape~\cite{chen2019text2shape} constructed many-to-many relations between language and 3D physical properties, enabling the generation control of color and shape. However, Text2Shape only collected 75K language descriptions for 15K chairs and tables. ShapeCraft~\cite{fu2022shapecrafter} gradually evolved more phrases, constructing a dataset with 369K shape-text pairs, named Text2Shape++. To support recursive generation, ShapeCraft captured local details with vector-quantized deep implicit functions. 
Recently, SDFusion~\cite{cheng2023sdfusion} proposed to embed conditional features to the denoising layer of diffusion training, allowing multi-modal input conditions.

However, restricted by available 3D data and corresponding captions, previous 3D native generative models can only handle limited categories. To support large-vocabulary 3D generation, pioneering works Point-E~\cite{nichol2022point} and Shap-E~\cite{jun2023shap} collected several millions of 3D assets and corresponding text captions. Point-E trained an image-to-point diffusion model, in which a CLIP visual latent code is fed into the transformer. Shap-E further introduced a latent projection to enable the reconstruction of SDF representation. Nonetheless, the proposed dataset is not released to the public. Instead, recent works have to conduct experiments based on a relatively smaller dataset Objaverse. LRM~\cite{hong2023lrm} proposed to learn an image-to-triplane latent space and then reshape the latent feature for reconstructing the triplane-based implicit representation. DMV3D~\cite{xu2023dmv3d} treated LRM as a denoising layer, further proposing a $T$-step diffusion model to generate high-quality results based on LRM. TextField3D~\cite{huang2023textfield3d} is proposed for open-vocabulary generation, where a text latent space is injected with dynamic noise to expand the expression range of latent features.

\subsection{Scene}

Early approach \cite{schwarz2020graf} utilize a Generative Adversarial Network (GAN) that explicitly incorporates a parametric function, known as radiance fields. This function takes the 3D coordinates and camera pose as input and generates corresponding density scalar and RGB values for each point in 3D space. However, GANs suffer from training pathologies including mode collapse and are difficult to train on data for which a canonical coordinate system does not exist, as is the case for 3D scenes \cite{bautista2022gaudi}. To overcome the problems, GAUDI \cite{bautista2022gaudi} learns a denoising diffusion model that is fit to a set of scene latents learned using an autodecoder. However, these models \cite{bautista2022gaudi} all have an inherent weakness of attempting to capture the entire scene into a single vector that conditions a neural radiance field. This limits the ability to fit complex scene distributions.
NeuralField-LDM \cite{kim2023neuralfield} first expresses the image and pose pairs as a latent code and learns the hierarchical diffusion model to complete the scene generation. However, the current method is time-consuming and resolution is relatively low. The recent $X^ 3$ employs a hierarchical voxel latent diffusion to generate a higher resolution 3D representation in a coarse-to-fine manner. 

\begin{figure*}[ht]
	\centering
	\includegraphics[width=\textwidth,trim=10bp 10bp 5bp 7bp,clip]{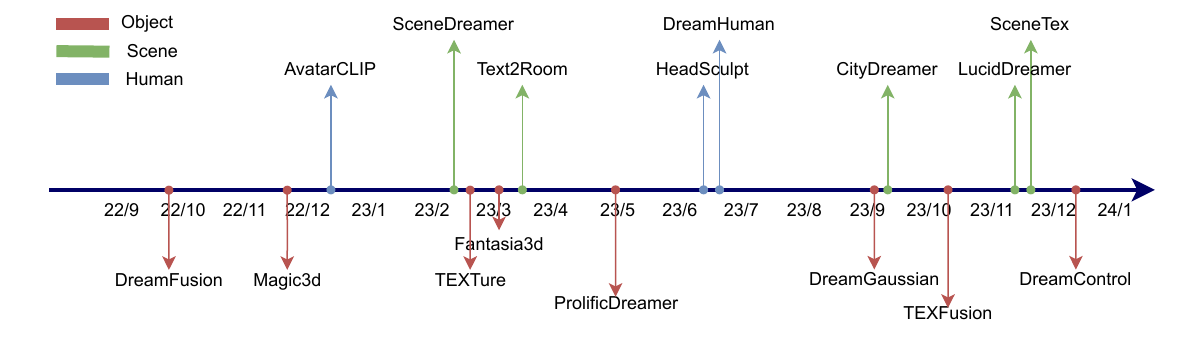} 
	\caption{Chronological overview of the most relevant 2D prior-based 3D generative methods.
	}
	\label{fig:2dprior}
\end{figure*}

\subsection{Human Avatar} \label{sec:native_human}

Early approaches of 3D human avatar generation rely on parametric models, which use a set of predefined parameters to create a 3D mesh of expressive human face or body.
3D morphable model (3DMM)~\cite{blanz2003face} is a statistical model that decomposes the intrinsic attributes of human face into identity, expression, and reflectance. These attributes are encoded as low-dimensional vectors and can be used to generate realistic 3D faces from 2D images or video footage.
For human bodies, one of the most widely used parametric models is the Skinned Multi-Person Linear (SMPL) model~\cite{loper2023smpl}, which uses a combination of linear and non-linear transformations to create a realistic 3D mesh of a human body. 
SMPL is based on a statistical body shape and pose model that was learned from a large dataset of body scans.
Despite the success of parametric models, they have several limitations, especially in modeling complex geometries such as hair and loose clothing.

Recent years have witnessed a significant shift towards learning-based methods for modeling 3D humans~\cite{chen2021towards}. 
These methods use deep learning algorithms to learn realistic and detailed human avatars from datasets of 3D scans or multi-view images.
PIFu~\cite{saito2019pifu} introduce pixel-aligned implicit function that can generate highly detailed 3D clothed humans with intricate shapes from a single image.
HeadNeRF~\cite{hong2022headnerf} proposes a NeRF-based parametric head model that can generate high-fidelity head images with the ability to manipulate rendering pose and various semantic attributes.
SMPLicit~\cite{corona2021smplicit} and gDNA~\cite{chen2022gdna} train 3D generative models of clothed humans using implicit functions from registered 3D scans. 
Recently, Rodin~\cite{wang2023rodin} presents a roll-out diffusion network based on tri-plane representation to learn detailed 3D head avatars from a large synthetic multi-view dataset.

\section{2D Prior-based 3D Generative Methods}

Previously, most 3D native generative methods were confined to constrained datasets like ShapeNet containing only fixed object categories. Recent advances in text-to-image diffusion models and gan offer new possibilities. Several milestone methods are presented in Figure~\ref{fig:2dprior}.

\subsection{Object}

DreamFusion~\cite{poole2023dreamfusion} pioneered the paradigm of optimizing a unique 3D representation per text input or per image, guided by powerful pretrained 2D diffusion models. This approach established a new foundation but also revealed key challenges ahead - namely, achieving high-fidelity quality in resolution, geometric detail and texture fidelity; ensuring consistent generation across diverse views, known as the "multi-face Janus problem"; and optimizing synthesis speed for interactive applications.  

To achieve high-fidelity quality, Magic3D~\cite{lin2023magic3d} introduced a coarse-to-fine optimization strategy with two stages. This approach improved both speed and quality. 
Fantasia3D~\cite{Chen_2023_ICCV} disentangled geometry and appearance modeling, advancing text-to-3D quality. For geometry, it relied on a hybrid scene representation and encoded extracted surface normals into the input of an image diffusion model. Regarding appearance, Fantasia3D introduced spatially-varying bidirectional reflectance distribution functions to learn surface materials for photorealistic rendering of generated geometry. While early methods suffered from over-saturation and low diversity issues, ProlificDreamer~\cite{wang2023prolificdreamer} introduced variational score distillation to address these challenges.  

However, due to Stable Diffusion's bias towards 2D front views, its 3D outputs tended to repeat front views from different angles rather than generating coherent 3D objects. In contrast to fine-tuning on multi-view 3D data to alleviate the Janus problem, some works explored alternative approaches - for example, DreamControl~\cite{huang2023dreamcontrol} utilized adaptive viewpoint sampling and boundary integrity metrics.

While previous per-sample optimization methods based on NeRF suffered from slow speeds for 3D generative tasks, the rapid development of 3DGS enabled a breakthrough. DreamGaussian ~\cite{tang2023dreamgaussian} incorporated 3DGS into generative 3D content creation, achieving around 10x acceleration compared to NeRF-based approaches. In contrast to the occupancy pruning utilized in NeRF, the progressive densification of 3D Gaussians converges significantly faster for these 3D generation problems. DreamGaussian introduced an efficient algorithm to convert the resulting Gaussians into textured meshes. This pioneering work demonstrated how 3DGS can enable much faster training for AIGC-3D.

In addition to joint geometry and texture generation, another paradigm involves texture mapping given predefined geometry, referred to as "texture painting" - also a form of content creation. Representative works in this area include TEXTure~\cite{TEXTure} and TexFusion~\cite{TexFusion}, which leverage pretrained depth-to-image diffusion models and apply iterative schemes to paint textures onto 3D models from multiple viewpoints. By disentangling texture generation from the separate challenge of geometric modeling, such approaches provide an alternative research direction worthy of exploration.

\subsection{Scene}

In most 2D prior-based scene generation approaches, the primary method is to utilize pretrained large models to generate partial scenes. Subsequently, an inpainting strategy is employed to generate large-scale scenes. Text2room \cite{hollein2023text2room} is a typical example of using 2D pretrained model to generate depth of 2D image. Then, the image is inpainted with more depths. These depths are merged to generate large scale scene. LucidDreamer \cite{chung2023luciddreamer} first generates multi-view consistent images from inputs by using inpaiting strategy. Then, the inpainted images are lifted to 3D space with estimated depth maps and aggregate the new depth maps into the 3D scene. 
SceneTex \cite{chen2023scenetex} generates scene textures for indoor scenes using depth-to-image diffusion priors. The core of this method lies in the proposal of a multiresolution texture field that implicitly encodes the appearance of the mesh. The target texture is then optimized using VSD loss in respective RGB renderings. Additionally, SceneDreamer~\cite{chen2023scenedreamer} introduces a Bird's Eye View (BEV) scene representation and a neural volumetric renderer. This framework learns an unconditional generative model from 2D image collections. CityDreamer~\cite{citydreamer}brings nature scene generation to the city level, enhancing the realism of city generation by decoupling buildings and background generation.

\subsection{Human Avatar}

In the field of text-guided 3D human generation, parametric models (see Section \ref{sec:native_human}) are extensively used as fundamental 3D priors, for they can provide accurate geometric initialization and reduce optimization difficulty considerably.
AvatarCLIP~\cite{hong2022avatarclip} is the first to combine vision-language models with implicit 3D representations derived from a parametric model to achieve zero-shot text-driven generation of full-body human avatars.
Following the success of generating 3D objects using SDS powered by pre-trained 2D latent diffusion models, recent works also extend such methods to human generation.
HeadSculpt~\cite{han2023headsculpt} generates consistent 3D head avatars by conditioning the pre-trained diffusion model on multi-view landmark maps obtained from a 3D parametric head model.

Following this scheme, DreamWaltz~\cite{huang2023dreamwaltz} proposes occlusion-aware SDS and skeleton conditioning to maintain 3D consistency and reduce artifacts during optimization.
By optimizing a NeRF in the semantic signed distance space of imGHUM with multiple fine-grained losses, 
DreamHuman~\cite{kolotouros2023dreamhuman} generates animatable 3D humans with instance-specific surface deformations.
HumanGaussian~\cite{liu2023humangaussian} incorporates SDS with SoTA 3DGS representation to achieve more efficient text-driven generation of 3D human avatars.

\begin{figure*}[ht]
	\centering
	\includegraphics[width=\textwidth,trim=10bp 10bp 5bp 7bp,clip]{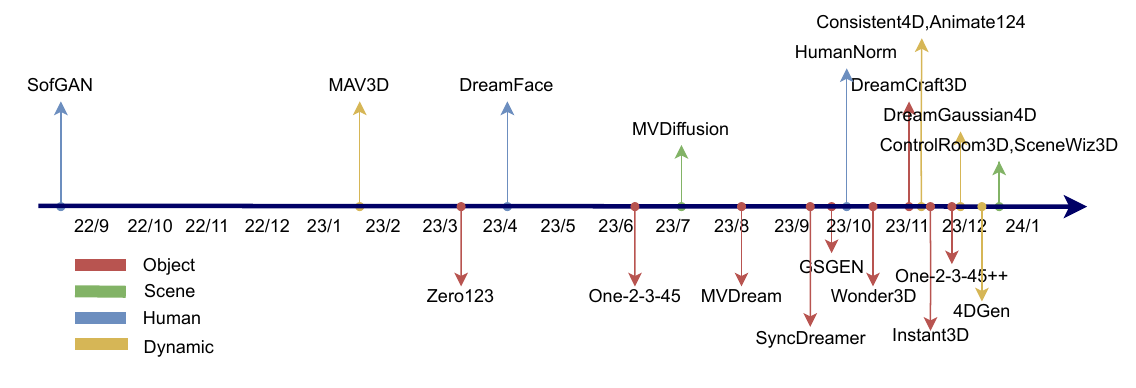}
	\caption{Chronological overview of the most relevant hybrid 3D generative methods.}
	\label{fig:hybrid}
\end{figure*}

\section{Hybrid 3D Generative Methods}

While early 3D native generative methods were limited by scarce 3D datasets, and 2D prior methods could only distill limited 3D geometric knowledge, researchers explored injecting 3D information into pretrained 2D models. Emerging approaches included fine-tuning Stable Diffusion on multi-view object images to generate consistent perspectives, as well as 3D reconstruction and generation from multiple views.

This paradigm shift addressed the above shortcomings, by leveraging both abundant 2D visual resources and targeted 3D supervision to move beyond the separate limitations of each individual approach. Several milestone methods are presented in Figure~\ref{fig:hybrid}.

\subsection{Object}
 
The first attempt is Zero123  ~\cite{liu2023zero}, which applies 3D data to fine-tune pre-trained 2D diffusion models, enabling the generation of novel views conditioned on a single input view. This insightful work demonstrated that Stable Diffusion inherently contained extensive 3D knowledge, which could be unlocked through multi-view fine-tuning. 
Building on this, One-2-3-45~\cite{liu2023one2345} leveraged Zero123 to produce multiple views. It then connected a reconstruction model, achieving 3D mesh generation from a single image in just 45 seconds with promising results. This approach moved past prior optimization relying on 2D priors, significantly increasing the speed of 3D generation.

While the newly generated views in Zero123 were consistent with the given view, consistency was not maintained between generated novel views. In response, several works aimed to produce multiple views simultaneously with inter-view consistency. SyncDreamer~\cite{liu2023syncdreamer}, MVDream ~\cite{shi2023mvdream} all enabled generating multiple perspectives at once, with information exchange between views to ensure consistency. Wonder3D ~\cite{long2023wonder3d} introduced a normal modal and fine-tuned a multi-view Stable Diffusion model to concurrently output RGB and normal maps across perspectives. One-2-3-45++  ~\cite{liu2023one2345plus} advanced multi-view 3D generation via an enhanced Zero123 module enabling simultaneous cross-view attention, alongside a multi-view conditioned 3D diffusion module performing coarse-to-fine textured mesh prediction over time.
UniDream~\cite{unidream} incorporates diffusion priors to generate multi-view albedo-normal information, which is further processed by a trained reconstruction model to produce 3D representations.

Several subsequent works introduced 3D prior initialization to improve quality of 3D generative content. 
Dreamcraft3d~\cite{sun2023dreamcraft3d} initialized a DMTet representation using score distillation sampling from a view-dependent diffusion model. Gsgen~\cite{chen2023textto3d} utilized Point-E to initialize 3D Gaussian positions for generation. By incorporating different forms of 3D structural information upfront, these papers produced more coherent 3D outputs compared to prior approaches lacking initialization techniques. 
 
Following the success of large-scale reconstruction models like LRM, Instant3d~\cite{li2023instant3dlrm} also utilized a two-stage approach. In the first stage, it performed multi-view generation. The second stage then directly regressed the NeRF from the generated images via a novel sparse-view reconstructor based on transformers. Combining multi-view Stable Diffusion and large-scale reconstruction models can effectively solve the problems of multi-face and generation speed.

\subsection{Scene}

There are recently several proposed methods on 3D scene generation.
MVDiffusion~\cite{tang2023mvdiffusion} simultaneously generates all images with global awareness, effectively addressing the common issue of error accumulation. The main feature of MVDiffusion is its ability to process perspective images in parallel using a pre-trained text-to-image diffusion model, while incorporating novel correspondence-aware attention layers to enhance cross-view interactions.
ControlRoom3D~\cite{schult2023controlroom3d} is a method to generate high-quality 3D room meshes with only a user-given textual description of the room style and a user-defined room layout. The naive layout-based 3D room generation method does not produce plausible meshes. To address the bad geometry problem and ensure consistent style, ControRoom3D leverages a guided panorama generation and geometry alignment modules.
SceneWiz3D~\cite{zhang2023scenewiz3d} introduces a method to synthesize high-fidelity 3D scenes from text. Given a text, a layout is first generated. Then, the Particle Swarm Optimization technique is applied to automatically place the 3D objects based on the layout and optimize the 3D scenes implicitly. The SceneWiz3D also leverage a RGBD panorama diffusion model to further improve the scene geometry.

\subsection{Human Avatar}

Several studies on 3D human generation have been leveraging both 2D and 3D datasets/priors to achieve more authentic and general synthesis of 3D humans, where 3D data provides accurate geometry and 2D data offers diverse appearance.
SofGAN~\cite{chen2022sofgan} proposes a controllable human face generator with a decoupled latent space of geometry and texture learned from unpaired datasets of 2D images and 3D facial scans. 
The 3D geometry is encoded to a semantic occupancy field to facilitate consistent free-viewpoint image generation.
Similarly, SCULPT~\cite{sanyal2023sculpt} also presents an unpaired learning procedure to effectively learn from medium-sized 3D scan datasets and large-scale 2D image datasets to learn disentangled distribution of geometry and texture of full-body clothed humans.
Get3DHuman~\cite{xiong2023get3dhuman} circumvent the requirement of 3D training data by combining two pre-trained networks, a StyleGAN-Human image generator and a 3D reconstructor.

Driven by the significant progress of recent text-to-image synthesis models, researchers have begun to use 3D human datasets to enhance the powerful 2D diffusion models to synthesize photorealistic 3D human avatars with high-frequency details.
DreamFace~\cite{zhang2023dreamface} generates photorealistic animatable 3D head avatars by bridging vision-language models with animatable and physically-based facial assets.
The realistic rendering quality is achieved by a dual-path appearance generation process, which combines a novel texture diffusion model trained on a carefully-collected physically-based texture dataset with the pre-trained diffusion prior.
HumanNorm~\cite{huang2023humannorm} proposes a two-stage diffusion pipeline for 3D human generation, which first generates detailed geometry by a normal-adapted diffusion model and then synthesizes photorealistic texture based on the generated geometry using a normal-aligned diffusion model. 
Both two diffusion models are fine-tuned on a dataset of 2.9K 3D human models.

\subsection{Dynamic}

Jointly optimized by 2D, 3D, as well as video prior, dynamic 3D generation is gaining significant attention recently. The pioneering work MAV3D~\cite{singer2023textto4d} proposed to generate a static 3D asset and then animate it with text-to-video diffusion, in which a 4D representation named hex-plane is introduced to expand 3D space with temporal dimension. Following MAV3D, a series of works created dynamic 3D content based on a static-to-dynamic pipeline, while different 4D representations and supervisions are proposed to improve generation quality. Animate124~\cite{zhao2023animate124} introduced an image-to-4D framework, in which hex-plane is replaced with 4D grid encoding. Except for static and dynamic stages, a refinement stage is further proposed to guide the semantic alignment of image input and 4D creation with ControlNet. 4D-fy~\cite{bahmani20234d} proposed a multi-resolution hash encoding that represents 3D and temporal space separately. It highlighted the importance of 3D generation quality and leveraged 3D prior to guide the optimization of the static stage. 

Recent works ~\cite{jiang2023consistent4d} attempted to reconstruct 3D scenes based on generated videos, introducing a new 4D pipeline that generates a video and then complements its 3D representation. 4DGen~\cite{yin20234dgen} made pseudo multi-view videos via multi-view diffusion prior and optimized the reconstruction of gaussian splattings based on multi-resolution hex-plane. DreamGaussian4d~\cite{ren2023dreamgaussian4d} deployed 3D-aware diffusion prior to supervise the multi-view reconstruction of a given video and refined the corresponding scene with video diffusion prior. 
\section{Future Direction}  

Despite the recent progress in 3D content generation, there are still many problems unsolved that will significantly impact the quality, efficiency and controllability of 3D conetent generation methods. In this section, we summarize these challenges and propose several future directions.

\subsection{Challenges}

In terms of quality, current AIGC-3D methods have some limitations. For geometry, they cannot generate compact meshes and fail to model reasonable wiring. For textures, they lack the ability to produce rich detail maps and it is difficult to eliminate the effects of lighting and shadows. Material properties are also not well supported. Regarding controllability, existing text/image/sketch to 3D approaches cannot precisely output 3D assets that meet conditional requirements. Editing capabilities are also insufficient. For speed, feed-forward and SDS methods based on GS are faster but offer lower quality than optimization approaches based on NeRF. Overall, generating 3D content at production-level quality, scale and precision remains unresolved.

\subsection{Data}

Regarding data, one challenge lies in collecting datasets containing billions of 3D objects, scenes and humans. This could potentially be achieved through an open-world 3D gaming platform, where users can freely create and upload their own custom 3D models.Additionally, it would be valuable to extract rich implicit 3D knowledge from multi-view images and videos. Large-scale datasets with such diverse, unlabeled 3D data hold great potential to advance unsupervised and self-supervised learning approaches for generative 3D content creation.

\subsection{Model}

There is a need to explore more effective 3D representations and model architectures capable of exhibiting scale-up performance alongside growing datasets. This presents an promising research avenue. Over the coming years, we may see the emergence of foundation models specialized for 3D content generatoin. Additionally, future large language models achieving high levels of multimodal intelligence, such as GPT-5/6, could theoretically understand images, text, and even programmatically operate 3D modeling software to an expert level. However, ensuring beneficial development of such powerful systems will require extensive research.

\subsection{Benchmark}

Currently, 3D content quality evaluation mainly relies on human ratings. ~\cite{wu2024gpt4vision} introduced an automated Human-Aligned Evaluator for text-to-3D generation. However, fully assessing 3D outputs is challenging since it requires comprehending both physical 3D properties and intended designs. Benchmarking 3D generation  has lagged progress in 2D image generation benchmarks. Developing robust metrics that holistically gauge geometric and textural fidelity based on photorealism standards could advance the field. 
\section{Conclusion}

In this survey, we have conducted a comprehensive analysis of 3D generative content techniques, encompassing 3D native generation, 2D prior-based 3D generation, and hybrid 3D generation. We have introduced a novel taxonomy to succinctly summarize the advancements made in recent methods for generating 3D content. Additionally, we have identified and summarized the unresolved challenges in this field, while also proposing several promising research directions. We firmly believe that this study will serve as an invaluable resource, guiding further advancements in the field as researchers tackle intriguing open problems by drawing inspiration from the ideas presented in this work.

{
\small
\bibliographystyle{ijcai24}
\bibliography{ijcai24}
}
\end{document}